# Learning distributions of shape trajectories from longitudinal datasets: a hierarchical model on a manifold of diffeomorphisms

Alexandre Bône      Olivier Colliot      Stanley Durrleman

The Alzheimer's Disease Neuroimaging Initiative

Institut du Cerveau et de la Moelle épinière, Inserm, CNRS, Sorbonne Université, Paris, France

Inria, Aramis project-team, Paris, France

`{alexandre.bone, olivier.colliot, stanley.durrleman}@icm-institute.org`

## Abstract

*We propose a method to learn a distribution of shape trajectories from longitudinal data, i.e. the collection of individual objects repeatedly observed at multiple time-points. The method allows to compute an average spatiotemporal trajectory of shape changes at the group level, and the individual variations of this trajectory both in terms of geometry and time dynamics. First, we formulate a non-linear mixed-effects statistical model as the combination of a generic statistical model for manifold-valued longitudinal data, a deformation model defining shape trajectories via the action of a finite-dimensional set of diffeomorphisms with a manifold structure, and an efficient numerical scheme to compute parallel transport on this manifold. Second, we introduce a MCMC-SAEM algorithm with a specific approach to shape sampling, an adaptive scheme for proposal variances, and a log-likelihood tempering strategy to estimate our model. Third, we validate our algorithm on 2D simulated data, and then estimate a scenario of alteration of the shape of the hippocampus 3D brain structure during the course of Alzheimer's disease. The method shows for instance that hippocampal atrophy progresses more quickly in female subjects, and occurs earlier in APOE4 mutation carriers. We finally illustrate the potential of our method for classifying pathological trajectories versus normal ageing.*

## 1. Introduction

### 1.1. Motivation

At the interface of geometry, statistics, and computer science, statistical shape analysis meets a growing number of applications in computer vision and medical image analysis. This research field has addressed two main statistical questions: atlas construction for cross-sectional shape datasets, and shape regression for shape time series. The former is the classical extension of a mean-variance analysis, which aims to estimate a mean shape and a covariance structure from observations of several individual instances of the same object or organ. The latter extends the concept of regression by estimating a spatiotemporal trajectory of shape changes from a series of observations of the same individual object at different time-points. The emergence of longitudinal shape datasets, which consist in the collection of individual objects repeatedly observed at multiple time-points, has raised the need for a combined approach. One needs statistical methods to estimate normative spatiotemporal models from series of individual observations which differ in shape and dynamics of shape changes across individuals. Such model should capture and disentangle the inter-subject variability in shape at each time-point and the temporal variability due to shifts in time or scalings in pace of shape changes. Considering individual series as samples along a trajectory of shape changes, this approach amounts to estimate a spatiotemporal distribution of trajectories, and has potential applications in various fields including silhouette tracking in videos, analysis of growth patterns in biology or modelling disease progression in medicine.

### 1.2. Related work

The central difficulty in shape analysis is that shape spaces are either defined by invariance properties [21, 40, 41] or by the conservation of topological properties [5, 8, 13, 20, 34], and therefore have intrinsically a structure of infinite dimensional Riemannian manifolds or Lie groups. Statistical Shape Models [9] are linear but require consistent points labelling across observations and have no topology preservation guarantees. A now usual approach is to use the action of a group of diffeomorphisms to define a metric on a shape space [29, 43]. This approach has been used to compute a "Fréchet mean" together with a covariance matrix in the tangent-space of the mean [1, 16, 17, 33, 44] from a cross-sectional dataset, and regression from time series of shape data [4, 15, 16, 18, 26, 32]. In [12, 14, 31], these tools have been used to estimate an average trajectory of shape changes from a longitudinal dataset using the convenient assumption that the parameters encoding inter-individual variability are independent of time. The work in [27] introduced the idea to use the parallel transport to translate the spatiotemporal patterns seen in one individual into the geometry of another one. The co-adjoint transport is used in [38] for the same purpose. Both estimate a group average trajectory from individual trajectories. The proposed models do not account for inter-individual variability in the time dynamics, which is of key importance in the absence of temporal markers of the progression to align the sequences. The same remarks can be applied to [6], which introduces a nice theoretical setting for spaces of trajectories, in the case of a fixed number of temporal observations across subjects. The need for temporal alignment in longitudinal data analysis is highlighted for instance in [13] with a diffeomorphism-based morphometry approach, or in [40, 41] with quotient manifolds. In [24, 37], a generative mixed-effects model for the statistical analysis of manifold-valued longitudinal data is introduced for the analysis of feature vectors. This model describes both the variability in the direction of the individual trajectories by introducing the concept of "exp-parallelization" which relies on parallel transport, and the pace at which those trajectories are followed using "time-warp" functions. Similar time-warps are used by the authors of [23] to refine their linear modeling approach [22].

### 1.3. Contributions

In this paper, we propose to extend the approach of [37] from low-dimensional feature vectors to shape data. Using an approach designed for manifold-valued data for shape spaces defined by the action of a group of diffeomorphisms raises several theoretical and computational difficulties. Are notably needed: a finite-dimensional set of diffeomorphisms with a Riemannian manifold structure; stable and efficient numerical schemes to compute Riemannian exponential and parallel transport operators on this manifold as no closed-form expressions are available; accelerated algorithmic convergence to cope with an hundreds of times larger dimensionality. To this end, we propose here:

- to formulate a generative non-linear mixed-effects model for a finite-dimensional set of diffeomorphisms defined by control points, to show that this set is stable under exp-parallelization, and to use an efficient numerical scheme for parallel transport;
- to introduce an adapted MCMC-SAEM algorithm with an adaptive block sampling of the latent variables, a specific sampling strategy for shape parameters based on random local displacements of the shape contours, and a vanishing tempering of the target log-likelihood;
- to validate our method on 2D simulated data and a large dataset of 3D brain structures in the context of Alzheimer's disease progression, and to illustrate the potential of our method for classifying spatiotemporal patterns, e.g. to discriminate pathological versus normal trajectories of ageing.

All in one, the proposed method estimates an average spatiotemporal trajectory of shape changes from a longitudinal dataset, together with distributions of space-shifts, time-shifts and acceleration factors describing the variability in shape, onset and pace of shape changes respectively.

## 2. Deformation model

### 2.1. The manifold of diffeomorphisms $\mathcal{D}_{c_0}$

We follow the approach taken in [12] built on the principles of the Large Deformation Diffeomorphic Metric Mapping (LDDMM) framework [30]. We note $d \in \{2, 3\}$ the dimension of the ambient space. We choose $k$ a Gaussian kernel of width $\sigma \in \mathbb{R}_+^*$ and $c$ a set of $n_{cp} \in \mathbb{N}$ "control" points $c = (c_1, ..., c_{n_{cp}})$ of the ambient space $\mathbb{R}^d$. For any set of "momentum" vectors $m = (m_1, ..., m_{n_{cn}})$, we define the "velocity" vector field $v : \mathbb{R}^d \to \mathbb{R}^d$ as the convolution $v(x) = \sum_{k=1}^{n_{cp}} k(c_k, x) m_k$ for any point $x$ of the ambient space $\mathbb{R}^d$. From initial sets of $n_{cp}$ control points $c_0$ and corresponding momenta $m_0$, we obtain the trajectories $t \to (c_t, m_t)$ by integrating the Hamiltonian equations:

$$\dot{c} = K_c \, m \quad ; \quad \dot{m} = -\frac{1}{2} \nabla_c \left\{ m^T K_c \, m \right\} \quad (1)$$

where $K_c$ is the $n_{cp} \times n_{cp}$ "kernel" matrix $[k(c_i, c_j)]_{ij}$, $\nabla(.)$ the gradient operator, and $(.)^T$ the matrix transposition.

Those trajectories prescribe the trajectory $t \to v_t$ in the space of velocity fields. The integration along such a path from the identity generates a flow of diffeomorphisms $t \to \phi_t$ of the ambient space [5]. We can now define:

$$\mathcal{D}_{c_0} = \big\{ \phi_1 \, ; \partial_t \phi_t = v_t \circ \phi_t, \, \phi_0 = \text{Id}, \, v_t = \text{Conv}(c_t, m_t)$$
$$(\dot{c}_t, \dot{m}_t) = \text{Ham}(c_t, m_t), \, m_0 \in \mathbb{R}^{dn_{cp}} \big\} \quad (2)$$

where $\text{Conv}(., .)$ and $\text{Ham}(., .)$ are compact notations for the convolution operator and the Hamiltonian equations (1) respectively. $\mathcal{D}_{c_0}$ has the structure of a manifold of finite dimension, where the metric at the tangent space $T_{\phi_1} \mathcal{D}_{c_0}$ is given by $K_{c_1}^{-1}$. It is shown in [29] that the proposed paths $t \to \phi_t$ are the paths of minimum deformation energy, and are therefore the geodesics of $\mathcal{D}_{c_0}$. These geodesics are fully determined by an initial set of momenta $m_0$.

Then, any point $x \in \mathbb{R}^d$ of the ambient space follows the trajectory $t \to \phi_t(x)$. Such trajectories are used to deform any point cloud or mesh embedded in the ambient space, defining a diffeomorphic deformation of the shape. Formally, this operation defines a shape space $\mathcal{S}_{c_0, y_0}$ as the orbit of a reference shape $y_0$ under the action of $\mathcal{D}_{c_0}$. The manifold of diffeomorphisms $\mathcal{D}_{c_0}$ is used as a proxy to manipulate shapes: all computations are performed in $\mathcal{D}_{c_0}$ or more concretely on a finite set of control points and momentum vectors, and applied back to the template shape $y_0$ to obtain a result in $\mathcal{S}_{c_0,y_0}$.

## 2.2. Riemmanian exponentials on $\mathcal{D}_{c_0}$

For any set of control points $c_0$, we define the exponential operator $\text{Exp}_{c_0} : m_0 \in \mathbb{R}^{dn_{cp}} \to \phi_1 \in \mathcal{D}_{c_0}$. Note that $\mathcal{D}_{c_0} = \{\text{Exp}_{c_0}(m_0) \mid m_0 \in \mathbb{R}^{dn_{cp}}\}$.

The following proposition ensures the stability of $\mathcal{D}_{c_0}$ by the exponential operator, i.e. that the control points obtained by applying successive compatible exponential maps with arbitrary momenta are reachable by an unique integration of the Hamiltonian equations from $c_0$:

**Proposition.** Let $c_0$ be a set of control points. $\forall \phi_1 \in \mathcal{D}_{c_0}$, $\forall w$ momenta, we have $\text{Exp}_{\phi_1(c_0)}(w) \in \mathcal{D}_{c_0}$.

**Proof.** We note $\phi_1' = \text{Exp}_{\phi_1(c_0)}(w) \in \mathcal{D}_{\phi_1(c_0)}$ and $c_1' = \phi_1' \circ \phi_1(c_0)$. By construction, there exist two paths $t \to \varphi_t$ in $\mathcal{D}_{c_0}$ and $s \to \varphi_s'$ in $\mathcal{D}_{\phi_1(c_0)}$ such that $\varphi_1' \circ \varphi_1(c_0) = c_1'$. Therefore there exist a diffeomorphic path $u \to \psi_u$ such that $\psi(c_0) = c_1'$. Concluding with [29], the path $u \to \psi_u$ of minimum energy exists, and is written $u \to \text{Exp}_{c_0}(u.m_0')$ for some $m_0' \in \mathbb{R}^{dn_{cp}}$. □

As a physical interpretation might be given to the integration time $t$ when building a statistical model, we introduce the notation $\text{Exp}_{c_0,t_0,t} : m_0 \in \mathbb{R}^{dn_{cp}} \to \phi_t \in \mathcal{D}_{c_0}$ where $\phi_t$ is obtained by integrating from $t = t_0$. Note that $\text{Exp}_{c_0} = \text{Exp}_{c_0,0,1}$.

On the considered manifold $\mathcal{D}_{c_0}$, computing exponentials – i.e. geodesics – therefore consists in integrating ordinary differential equations. This operation is direct and computationally tractable. The top line on Figure 1 plots a geodesic $\gamma : t \to \phi_t$ applied to the top-left shape $y_0$.

## 2.3. Parallel transport and exp-parallels on $\mathcal{D}_{c_0}$

In [37] is introduced the exp-parallelism, which is a generalization of Euclidian parallelism to geodesically-complete manifolds. It relies on the Riemannian parallel transport operator, which we propose to compute using the fanning scheme [28]. This numerical scheme only requires the exponential operator to approximate the parallel transport along a geodesic, with proved convergence.

We note $P_{c_0,m_0,t_0,t} : \mathbb{R}^{dn_{cp}} \to \mathbb{R}^{dn_{cp}}$ the parallel transport operator, which transports any momenta $w$ along the geodesic $\gamma : t \to \phi_t = \text{Exp}_{c_0,t_0,t}(m_0)$ from $\phi_{t_0}$ to $\phi_t$. For any $c_0, m_0, w$ and $t_0$, we can now define the curve:

$$t \to \eta_{c_0,m_0,t_0,t}(w) = \text{Exp}_{\gamma(t)(c_0)}\left[P_{c_0,m_0,t_0,t}(w)\right]. \quad (3)$$

This curve, that we will call exp-parallel to $\gamma$, is well-defined on the manifold $\mathcal{D}_{c_0}$, according to the proposition of Section 2.2. Figure 1 illustrates the whole procedure. From the top-left shape, the computational scheme is as follow: integrate the Hamiltonian equations to obtain the control points $c(t)$ (red crosses) and momenta $m(t)$ (bold blue arrows); compute the associated velocity fields $v_t$ (light blue arrows); compute the flow $\gamma : t \to \phi_t$ (shape progression); transport the momenta $w$ along $\gamma$ (red arrows); compute the exp-parallel curve $\eta$ by repeating the three first steps along the transported momenta.

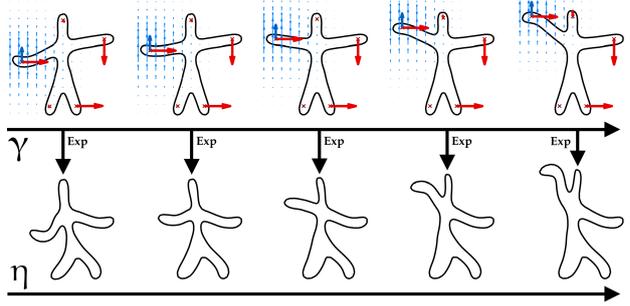

Figure 1: Samples from a geodesic $\gamma$ (top) and an exp-parallelized curve $\eta$ (bottom) on $\mathcal{S}_{c_0,y_0}$. Parameters encoding the geodesic are the blue momenta attached to control points and plotted together with the associated velocity fields. Momenta in red are parallel transported along the geodesic and define a deformation mapping each frame of the geodesic to a frame of $\eta$. Exp-parallelization allows to transport a shape trajectory from one geometry to another.

## 3. Statistical model

For each individual $1 \leq i \leq N$ are available $n_i$ longitudinal shape measurements $y = (y_{i,j})_{1 \leq j \leq n_i}$ and associated times $(t_{i,j})_{1 \leq j \leq n_i}$.

### 3.1. The generative statistical model

Let $c_0$ a set of control points and $m_0$ associated momenta. We call $\gamma$ the geodesic $t \to \text{Exp}_{c_0,t_0,t}(m_0)$ of $\mathcal{D}_{c_0}$. Let $y_0$ be a template mesh shape embedded in the ambient space. For a subject $i$, the observed longitudinal shape measurements $y_{i,0}, ..., y_{i,n_i}$ are modeled as sample points at times $\psi_i(t_{i,j})$ of an exp-parallel curve $t \to \eta_{c_0,m_0,t_0,t}(w_i)$ to this geodesic $\gamma$, plus additional noise $\epsilon_{i,j}$:

$$y_{i,j} = \eta_{c_0,m_0,t_0,\psi_i(t_{i,j})}(w_i) \circ y_0 + \epsilon_{i,j}. \quad (4)$$

The time warp function $\psi_i$ and the space-shift momenta $w_i$ respectively encode for the individual time and space variability. The time-warp is defined as an affine reparametrization of the reference time $t$: $\psi_i(t) = \alpha_i(t - t_0 - \tau_i) + t_0$ where the individual time-shift $\tau_i \in \mathbb{R}$ allows an inter-individual variability in the stage of evolution, and the individual acceleration factor $\alpha_i \in \mathbb{R}_+^*$ a variability in the pace of evolution. For convenience, we write $\alpha_i = \exp(\xi_i)$. In the spirit of Independent Component Analysis [19], the space-shift momenta $w_i$ is modeled as the linear combination of $n_s$ sources, gathered in the $n_{cp} \times n_s$ matrix $A$:

$w_i = A_{m_0^\perp} s_i$. Before computing this superposition, each column $c_l(A)$ of A has been projected on the hyperplane $m_0^\perp$ for the metric $K_{c_0}$, ensuring the orthogonality between $m_0$ and $w_i$. As argued in [37], this orthogonality is fundamental for the identifiability of the model. Without this constraint, the projection of the space shifts $(w_i)_i$ on $m_0$ could be confounded with the acceleration factors $(\alpha_i)_i$.

## 3.2. Mixed-effects formulation

We use either the current [42] or varifold [7] noise model for the residuals $\epsilon_{i,j}$, allowing our method to work with input meshes *without any point correspondence*. In this setting, we note $\epsilon_{i,j} \stackrel{iid}{\sim} \mathcal{N}(0, \sigma_\epsilon^2)$. The other previously introduced variables are modeled as random effects $z$, with: $y_0 \sim \mathcal{N}(\overline{y_0}, \sigma_y^2)$, $c_0 \sim \mathcal{N}(\overline{c_0}, \sigma_c^2)$, $m_0 \sim \mathcal{N}(\overline{m_0}, \sigma_m^2)$, $A \sim \mathcal{N}(\overline{A}, \sigma_A^2)$, $\tau_i \stackrel{iid}{\sim} \mathcal{N}(0, \sigma_\tau^2)$, $\xi_i \stackrel{iid}{\sim} \mathcal{N}(0, \sigma_\xi^2)$, $s_i \stackrel{iid}{\sim} \mathcal{N}(0, 1)$. We define $\theta = (\overline{y_0}, \overline{c_0}, \overline{m_0}, \overline{A}, t_0, \sigma_\tau^2, \sigma_\xi^2, \sigma_\epsilon^2)$ the fixed effects i.e. the parameters of the model. The remaining variance parameters $\sigma_y^2, \sigma_c^2, \sigma_m^2$ and $\sigma_A^2$ can be chosen arbitrarily small. Standard conjugate distributions are chosen as Bayesian priors on the model parameters: $\overline{y_0} \sim \mathcal{N}(\overline{\overline{y_0}}, \varsigma_\mathcal{O}^2)$, $\overline{c_0} \sim \mathcal{N}(\overline{\overline{c_0}}, \varsigma_c^2)$, $\overline{m_0} \sim \mathcal{N}(\overline{\overline{m_0}}, \varsigma_m^2)$, $\overline{A} \sim \mathcal{N}(\overline{\overline{A}}, \varsigma_A^2)$, $t_0 \sim \mathcal{N}(\overline{\overline{t_0}}, \varsigma_t^2)$, $\sigma_\tau^2 \sim \mathcal{IG}(m_\tau, \sigma_{\tau,0}^2)$, $\sigma_\xi^2 \sim \mathcal{IG}(m_\xi, \sigma_{\xi,0}^2)$, and $\sigma_\epsilon^2 \sim \mathcal{IG}(m_\epsilon, \sigma_{\epsilon,0}^2)$ with inverse-gamma distributions on variance parameters. Those priors ensure the existence of the maximum a posteriori (MAP) estimator. In practice, they regularize and guide the estimation procedure.

The proposed model belongs to the curved exponential family (see supplementary material, which gives the complete log-likelihood). In this setting, the algorithm introduced in the following section has a proved convergence.

We have defined a distribution of trajectories that could be noted $t \to y(t) = f_{\theta,t}(z)$ where $z$ is a random variable following a normal distribution. We call $t \to f_{\theta,t}(E[z])$ the *average* trajectory, which may not be equal to the *expected* trajectory $t \to E[f_{\theta,t}(z)]$ in the general non-linear case.

## 4. Estimation

### 4.1. The MCMC-SAEM algorithm

The Expectation Maximization (EM) algorithm [11] allows to estimate the parameters of a mixed-effects model with latent variables, here the random effects $z$. It alternates between an expectation (E) step and a maximization (M) one. The E step is intractable in our case, due to the non-linearity of the model. In [10] is introduced and proved a stochastic approximation of the EM algorithm, where the E step is replaced by a simulation (S) step followed by an approximation (A) one. The S step requires to sample $q(z|y, \theta^k)$, which is also intractable in our case. In the case of curved exponential models, the authors in [2] show that the convergence holds if the S step is replaced by a single transition of an ergodic Monte-Carlo Markov Chain (MCMC) whose stationary distribution is $q(z|y, \theta^k)$. This global algorithm is called the Monte-Carlo Markov Chain Stochastic Approximation Expectation-Maximization (MCMC-SAEM), and is exploited in this paper to compute the MAP estimate of the model parameters $\theta_{\text{map}} = \max_\theta \int q(y, z|\theta)\, \mathrm{d}z$.

### 4.2. The adaptative block sampler

We use a block formulation of the Metropolis-Hasting within Gibbs (MHwG) sampler in the S-MCMC step. The latent variables $z$ are decomposed into $n_b$ natural blocks: $z = \{y_0, c_0, m_0, [c_l(A)]_l, [\tau_i, \xi_i, s_i]_i\}$. Those blocks have highly heterogeneous sizes, e.g. a single scalar for $\tau_i$ versus possibly thousands for $y_0$, for which we introduce a specific proposal distribution in Section 4.3.

For all the other blocks, we use a symmetric random walk MHwG sampler with normal proposal distributions of the form $\mathcal{N}(0, \sigma_b^2 \text{Id})$ to perturb the current block state $z_b^k$. In order to achieve reasonable acceptance rates $ar$ i.e. around $ar^\star = 30\%$ [36], the proposal standard deviations $\sigma_b$ are dynamically adapted every $n_{\text{adapt}}$ iterations by measuring the mean acceptance rates $\overline{ar}$ over the last $n_{\text{detect}}$ iterations, and applying, for any $b$:

$$\sigma_b \leftarrow \sigma_b + \frac{1}{k^\delta} \frac{\overline{ar} - ar^\star}{(1 - ar^\star)\mathbb{1}_{ar \geq ar^\star} + ar^\star \mathbb{1}_{ar < ar^\star}} \quad (5)$$

with $\delta > 0.5$. Inspired by [3], this dynamic adaptation is performed with a geometrically decreasing step-size $k^{-\delta}$, ensuring the vanishing property of the adaptation scheme and the convergence of the whole algorithm [2, 3]. It proved very efficient in practice with $n_{\text{adapt}} = n_{\text{detect}} = 10$ and $\delta = 0.51$, for any kind of data.

### 4.3. Efficient sampling of smooth template shapes

The first block $z_1 = y_0$ i.e. the coordinates of the points of the template mesh, is of very high dimension: naively sampling over each scalar value of its numerical description would result both in unnatural distorted shapes and a daunting computational burden.

We propose to take advantage of the geometrical nature of $y_0$ and leverage the framework introduced in Section 2 by perturbing the current block state $z_1^k$ with a small displacement field $v$, obtained by the convolution of random momenta on a pre-selected set of control points. This proposal distribution can be seen as a normal distribution $\mathcal{N}(0, \sigma_1^2 D^T D)$ where $\sigma_1^2$ is the variance associated with the random momenta, and $D$ the convolution matrix. In practice, dynamically adapting the proposal variance $\sigma_1^2$ and selecting regularly-spaced shape points as control points proved efficient.

### 4.4. Tempering

The MCMC-SAEM is proved convergent toward a local maximum of $\theta \to \int q(y, z|\theta) \, dz$. In practice, the dimensionality of the energetic landscape $q(y, z|\theta)$ and the presence of multiple local maxima can make the estimation procedure sensitive to initialization. Inspired by the globally-convergent *simulated annealing* algorithm, [25] proposes to carry out the optimization procedure in a smoothed version of the original landscape $q_T(y, z|\theta)$. The temperature parameter $T$ controls this smoothing, and should decrease from large values to 1, for which $q_T = q$.

We propose to introduce such a temperature parameter only for the population variables $z_{\text{pop}}$. The tempered version of the complete log-likelihood is given as supplementary material. In our experiments, the chosen temperature sequence $T^k$ remains constant at first, and then geometrically decreases to unity. Implementing this "tempering" feature had a dramatic impact on the required number of iterations before convergence, and greatly improved the robustness of the whole procedure. Note that the theoretical convergence properties of the MCMC-SAEM are not degraded, since the tempered phase of the algorithm can be seen as an initializing heuristic, and may actually be improved.

---

**Algorithm 1:** Estimation of the longitudinal deformations model with the MCMC-SAEM.
Code publicly available at: www.deformetrica.org.

**input** : Longitudinal dataset of shapes $y = (y_{i,j})_{i,j}$. Initial parameters $\theta^0$ and latent variables $z^0$. Geometrically decreasing sequence of step-sizes $\rho^k$.
**output:** Estimation of $\theta_{\text{map}}$. Samples $(z^s)_s$ approximately distributed following $q(z \mid y, \theta_{\text{map}})$.
*Initialization*: set $k = 0$ and $S^0 = S(z^0)$.
**repeat**
  *Simulation*: **foreach** block of latent variables $z_b$ **do**
    Draw a candidate $z_b^c \sim p_b(.|z_b^k)$.
    Set $z^c = (z_1^{k+1}, ..., z_{b-1}^{k+1}, z_b^c, z_{b+1}^k, ..., z_{n_b}^k)$.
    Compute the geodesic $\gamma: t \to \text{Exp}_{c_0, t_0, t}(m_0)$.
    $\forall i$, compute $w_i = A_{m_0}^\perp s_i$.
    $\forall i$, compute $w: t \to P_{c_0, m_0, t_0, t}(w_i)$.
    $\forall i, j$, compute $\text{Exp}_{\gamma[\psi_i(t_{i,j})](c_0)}\{w[\psi_i(t_{i,j})]\}$.
    Compute the acceptation ratio $\omega = \min\left[1, \frac{q(z^c|y,\theta^k)}{q(z^k|y,\theta^k)}\right]$.
    **if** $u \sim \mathcal{U}(0, 1) < \omega$ **then** $z_b^{k+1} \leftarrow z_b^c$ **else** $z_b^{k+1} \leftarrow z_b^k$.
  **end**
  *Stochastic approx.*: $S^{k+1} \leftarrow S^k + \rho^k [S(z^{k+1}) - S^k]$.
  *Maximization*: $\theta^{k+1} \leftarrow \theta^\star(S^{k+1})$.
  *Adaptation*: **if** remainder$(k+1, n_{\text{adapt}}) = 0$ **then** update the proposal variances $(\sigma_b^2)_b$ with equation (5).
  *Increment*: set $k \leftarrow k + 1$.
**until** convergence;

---

### 4.5. Sufficient statistics and maximization step

Exhibiting the sufficient statistics $S_1 = y_0$, $S_2 = c_0$, $S_3 = m_0$, $S_4 = A$, $S_5 = \sum_i t_0 + \tau_i$, $S_6 = \sum_i (t_0 + \tau_i)^2$, $S_7 = \sum_i \xi_i^2$ and $S_8 = \sum_i \sum_j \|y_{i,j} - \eta_{c_0, m_0, t_0, \psi_i(t_{i,j})}(w_i) \circ y_0\|^2$, the update of the model parameters $\theta \leftarrow \theta^\star$ in the M step of the MCMC-SAEM can be derived in closed-form. The explicit expressions are given as supplementary material.

## 5. Experiments

### 5.1. Validation with simulated data in $\mathbb{R}^2$

**Convergence study.** To validate the estimation procedure, we first generate synthetic data directly from the model without additional noise. Our choice of reference geodesic $\gamma$ is plotted on top line of the previously introduced Figure 1: the template $y_0$ is the top central shape, the chosen five control points $c_0$ are the red crosses, and the momenta $m_0$ the bold blue arrow. Those parameters are completed with $t_0 = 70$, $\sigma_\tau = 1$, $\sigma_\xi = 0.1$. With $n_s = 4$ independent components, we simulate $N = 100$ individual trajectories and sample $\langle n_i \rangle_i = 5$ observations from each.

The algorithm is run ten times. Figure 2 plots the evolution of the error on the parameters along the estimation procedure in log scale. Each color corresponds to a different run: the algorithm converges to the same point each time, as it is confirmed by the small variances on the residual errors indicated in Table 1. Those residual errors come from the finite number of observations of the generated dataset and the Bayesian priors, but are satisfyingly small, as qualitatively confirmed by Figure 3. The estimated mean trajectory, in bold, matches the true one, given by the top line of Figure 1. Figure 3 also illustrates the ability of our method to reconstruct continuous individual trajectories.

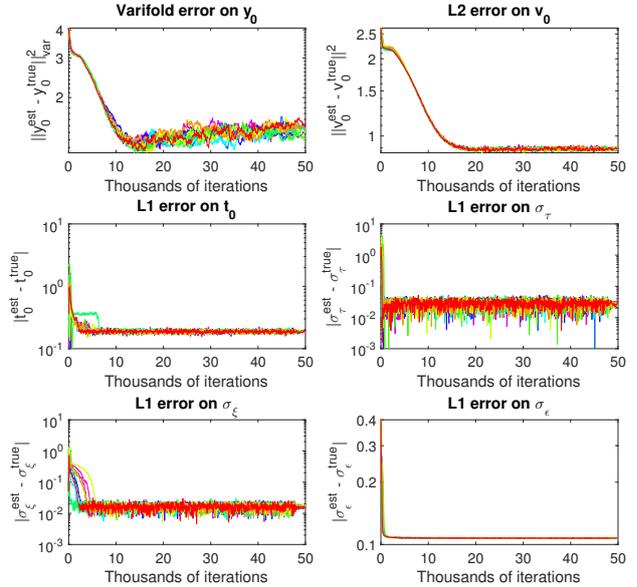

Figure 2: Error on the population parameters along the estimation procedure, with logarithmic scales. The residual on the template shape $y_0$ is computed with the varifold metric.

| $\lVert\Delta y_0\rVert^2_{\text{var.}}$ | $\lVert\Delta v_0\rVert^2$ | $\lvert\Delta t_0\rvert$ | $\lvert\Delta\sigma_\tau\rvert$ | $\lvert\Delta\sigma_\xi\rvert$ | $\lvert\Delta\sigma_\epsilon\rvert$ | $\langle\lVert\Delta v_i\rVert^2\rangle_i$ | $\langle\lvert\Delta\xi_i\rvert\rangle_i$ | $\langle\lvert\Delta\tau_i\rvert\rangle_i$ |
|---|---|---|---|---|---|---|---|---|
| $1.43_{\pm 5.6\%}$ | $0.89_{\pm 0.7\%}$ | $0.19_{\pm 2.7\%}$ | $0.029_{\pm 13.2\%}$ | $0.017_{\pm 7.6\%}$ | $0.11_{\pm 0.1\%}$ | $2.47_{\pm 1.7\%}$ | $0.022_{\pm 6.7\%}$ | $0.19_{\pm 0.8\%}$ |

Table 1: Absolute residual errors on the estimated parameters and associated relative standard deviations across the 10 runs. Are noted $v_0 = \text{Conv}(c_0, m_0)$ and $v_i = \text{Conv}(c_0, w_i)$. The operator $\langle.\rangle_i$ indicates an average over the index $i$. Residuals are satisfyingly small, as it can be seen for $\lvert\Delta t_0\rvert$ for instance when compared with the time-span $\max\lvert\Delta t_{ij}\rvert = 4$. The low standard deviations suggest that the stochastic estimation procedure is stable and reproduces very similar results at each run.

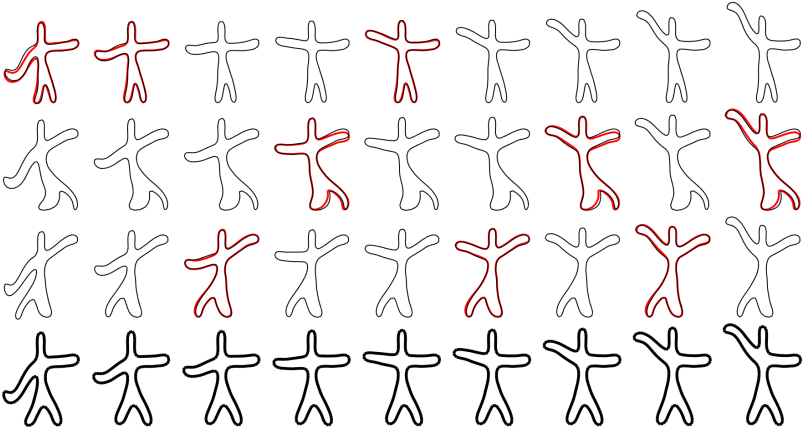

Figure 3: Estimated mean progression (bottom line in bold), and three reconstructed individual scenarii (top lines). Input data is plotted in red in the relevant frames, demonstrating the reconstruction ability of the estimated model. Our method is able to disentangle the variability in shape, starting time of the arm movement and speed.

**Personalizing the model to unseen data.** Once a model has been learned i.e. the parameters $\theta_{\text{map}}$ have been estimated, it can easily be personalized to the observations $y_{\text{new}}$ of a new subject by maximizing $q(y_{\text{new}}, z_{\text{new}} \mid \theta_{\text{map}})$ for the low-dimensional latent variables $z_{\text{new}}$. We implemented this maximization procedure with the Powell's method [35], and evaluated it by registering the simulated trajectories to the true model. Table 2 gathers the results for the previously-introduced dataset with $\langle n_i\rangle_i = 5$ observations per subject, and extended ones with $\langle n_i\rangle_i = 7$ and $9$. The parameters are satisfyingly estimated in all configurations: the reconstruction error measured by $\lvert\Delta\sigma_\epsilon\rvert$ remains as low as in the previous experiment (see Table 1, Figure 3). The acceleration factor is the most difficult parameter to estimate with small observation windows of the individual trajectories; at least two observations are needed to obtain a good estimate.

## 5.2. Hippocampal atrophy in Alzheimer's disease

**Longitudinal deformations model on MCIc subjects.** We extract the T1-weighted magnetic resonance imaging measurements of $N = 100$ subjects from the ADNI database, with $\langle n_i\rangle_i = 7.6$ datapoints on average. Those subjects present mild cognitive impairments, and are eventually diagnosed with Alzheimer's disease (MCI converters, noted MCIc). In a pre-processing phase, the 3D images are affinely aligned and the segmentations of the right-hemisphere hippocampi are transformed into a surface meshes. Each affine transformation is then applied to the corresponding mesh, before rigid alignment of follow-up meshes on the baseline one. The hippocampus is a subcortical brain structure which plays a central role in memory, and experiences atrophy during the development of Alzheimer's disease. We initialize the geodesic population parameters $y_0, c_0, m_0$ with a geodesic regression [15, 16] performed on a single subject. The reference time $t_0$ is initialized to the mean of the observation times $(t_{i,j})_{i,j}$ and $\sigma_\tau^2$ to the corresponding variance. We choose to estimate $n_s = 4$ independent components and initialize the corresponding matrix $A$ to zero, so as the individual latent variables $\tau_i, \xi_i, s_i$. After 10,000 iterations, the parameter estimates stabilized.

| Experience | $\lvert\Delta\sigma_\epsilon\rvert$ | $\langle\lvert\Delta s_i\rvert\rangle_i$ | $\langle\lvert\Delta\xi_i\rvert\rangle_i$ | $\langle\lvert\Delta\tau_i\rvert\rangle_i$ |
|---|---|---|---|---|
| $\langle n_i\rangle_i = 5$ | 0.110 | 3.34% | 37.0% | 5.45% |
| $\langle n_i\rangle_i = 7$ | 0.095 | 2.98% | 16.2% | 3.86% |
| $\langle n_i\rangle_i = 9$ | 0.087 | 2.38% | 11.9% | 3.28% |

Table 2: Residual errors metrics for the longitudinal registration procedure, for three simulated datasets. The absolute residual error on $\sigma_\epsilon$ is given, the other errors are given in percentage of the simulation standard deviation.

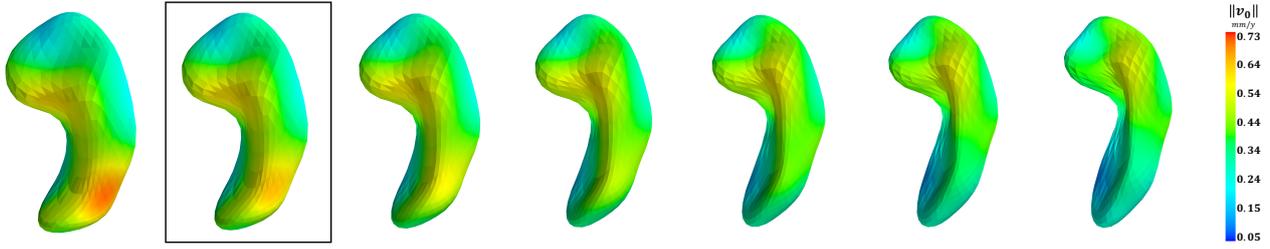

Figure 4: Estimated mean progression of the right hippocampi. Successive ages: 69.3y, 71.8y (i.e. the template $y_0$), 74.3y, 76.8y, 79.3y, 81.8y, and 84.3y. The color map gives the norm of the velocity field $\|v_0\|$ on the meshes.

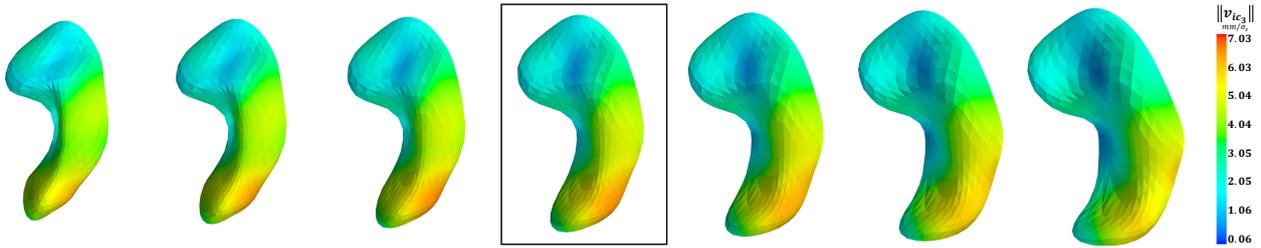

Figure 5: Third independent component. The plotted hippocampi correspond to $s_{i,3}$ successively equal to: -3, -2, -1, 0 (i.e. the template $y_0$), 1, 2 and 3. Note that this component is orthogonal to the temporal trajectory displayed in Figure 4.

Figure 4 plots the estimated mean progression, which exhibits a complex spatiotemporal atrophy pattern during disease progression: a pinching effect at the "joint" between the head and the body, combined with a specific atrophy of the medial part of the tail. Figure 5 plots an independent component, which is orthogonal to the mean progression by construction. This component seems to account for the inter-subject variability in the relative size of the hippocampus head compared to its tail.

We further examine the correlation between individual parameters and several patients characteristics. Figure 6 exhibits the strong correlation between the estimated individual time-shifts $\tau_i$ and the age of diagnostic $t_i^{\text{diag}}$, suggesting that the hippocampal atrophy correlates well with the cognitive symptoms. The few outliers above the regression line might have resisted better to the atrophy of their hippocampus with a higher brain plasticity, in line with the cognitive reserve theory [39]. The few outliers below this line could have developed a subform of the disease, with delayed atrophy of their hippocampi. Further investigation is required to rule out potential convergence issues in the optimization procedure. Figures 7, 8, 9 propose group comparisons based on the estimated individual parameters: the acceleration factor $\alpha_i$, time-shift $\tau_i$ and space-shift $s_{i,3}$ in the direction of the third component (see Figure 5). The distributions of those parameters are significantly different for the Mann-Whitney statistical test when dividing the $N = 100$ MCIc subjects according to gender, APOE4 mutation status, and onset age $t_0 + \tau_i$ respectively.

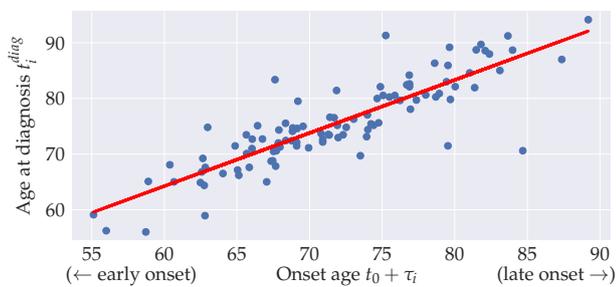

Figure 6: Comparison of the estimated individual time-shifts $\tau_i$ versus the age of diagnostic $t_i^{\text{diag}}$. $R^2 = 0.74$.

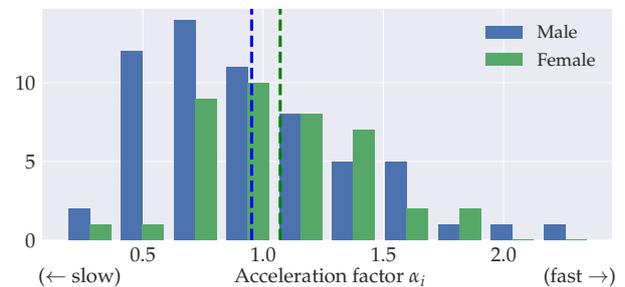

Figure 7: Distributions of acceleration factors $\alpha_i$ according to the gender. Hippocampal atrophy is faster in female subjects ($p = 0.045$).

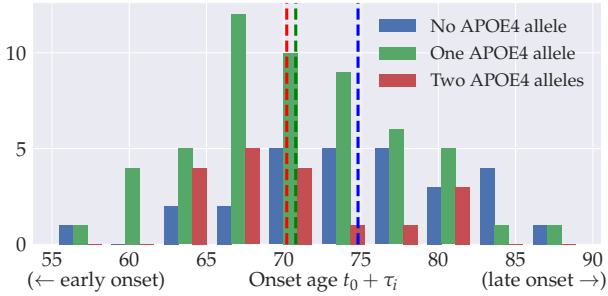

Figure 8: Distributions of time-shifts $\tau_i$ according to the number of APOE4 alleles. Hippocampal atrophy occurs earlier in carriers of 1 or 2 alleles ($p = 0.017$ and $0.015$).

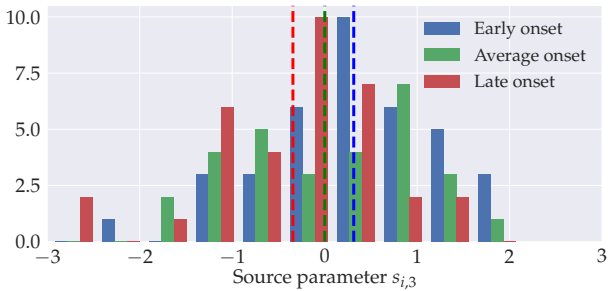

Figure 9: Distribution of the third source term $s_{i,3}$ according to the categories $\{\tau_i \leq -3\}$, $\{-3 < \tau_i < 3\}$, and $\{3 \leq \tau_i\}$. Hippocampal atrophy seems to occurs later in subjects presenting a lower volume ratio of the hippocampus tail over the hippocampus head ($p = 0.0049$).

| $\langle n_i \rangle_i$ | Shape features | Naive feature | All features |
| --- | --- | --- | --- |
| 1 | 71% ±4.5 [lr] | 50% ±5.0 [nb] | 58% ±5.0 [lr] |
| 2 | 77% ±4.3 [lr] | 58% ±4.9 [5nn] | 68% ±4.7 [dt] |
| 4 | 79% ±4.1 [svm] | 67% ±4.7 [5nn] | 80% ±4.0 [lr] |
| 5 | 77% ±4.2 [nn] | 77% ±4.2 [lr] | 82% ±3.8 [nb] |
| 6.86 | 83% ±3.7 [lr] | 80% ±4.0 [lr] | 86% ±3.4 [lr] |

Table 3: Mean classification scores and associated standard deviations, computed on 10,000 bootstrap samples from the test dataset. Across all tested classifiers (`sklearn` default hyperparameters), only the best performing one is reported in each case: [lr] logistic regression, [nb] naive Bayes, [5nn] 5 nearest neighbors, [dt] decision tree, [nn] neural network, [svm] linear support vector machine.

**Classifying pathological trajectories vs. normal ageing.** We processed another hundred of individuals from the ADNI database ($\langle n_i \rangle_i = 7.37$), choosing this time control subjects (CN). We form two balanced datasets, each containing 50 MCIc and 50 CN. We learn two distinct longitudinal deformations model on the training MCIc ($N = 50$, $\langle n_i \rangle_i = 8.14$) and CN ($N = 50$, $\langle n_i \rangle_i = 8.08$) subjects. We personalize both models to all the 200 subjects, and use the scaled and normalized differences $z_i^{\text{MCIc}} - z_i^{\text{CN}}$ as feature vectors of dimension 6, on which a list of standard classifiers are trained and tested to predict the label in $\{\text{MCIc}, \text{CN}\}$. For several number of observations per test subject $\langle n_i \rangle_i$ configurations, we compute confidence intervals by bootstraping the test set. Table 3 compares the results with a naive approach, using as unique feature the slope of individually-fitted linear regressions of the hippocampal volume with age. Classifiers performed consistently better with the features extracted from the longitudinal deformations model, even with a single observation. The classification performance increases with the number of available observations per subject. Interestingly, from $\langle n_i \rangle_i = 4$ pooling the shape and volume features yields an improved performance, suggesting complementarity.

## 6. Conclusion

We proposed a hierarchical model on a manifold of diffeomorphisms estimating the spatiotemporal distribution of longitudinal shape data. The observed shape trajectories are represented as individual variations of a group-average, which can be seen as the mean progression of the population. Both spatial and temporal variability are estimated directly from the data, allowing the use of unaligned sequences. This feature is key for applications where no objective temporal markers are available, as it is the case for Alzheimer's disease progression for instance, whose onset age and pace of progression vary among individuals. Our model builds on the principles of a generic longitudinal modeling for manifold-valued data [37]. We provided a coherent theoretical framework for its application to shape data, along with the needed algorithmic solutions for parallel transport and sampling on our specific manifold. We estimated our model with the MCMC-SAEM algorithm both with simulated and real data. The simulation experiments confirmed the ability of the proposed algorithm to retrieve the optimal parameters in realistic scenarii. The application to medical imaging data, namely segmented hippocampi brain structures of Alzheimer's diseased patients, delivered results coherent with medical knowledge, and provides more detailed insights into the complex atrophy pattern of the hippocampus and its variability across patients. In future work, the proposed method will be leveraged for automatic diagnosis and prognosis purposes. Further investigations are also needed to evaluate the algorithm convergence with respect to the number of individual samples.

**Acknowledgments.** This work has been partly funded by the European Research Council with grant 678304, European Union's Horizon 2020 research and innovation program with grant 666992, and the program Investissements d'avenir ANR-10-IAIHU-06.

## Appendix: supplementary material

We introduce the onset age individual random variable $t_i = t_0 + \tau_i \sim \mathcal{N}(t_0, \sigma_\tau^2)$ instead of the time shift $\tau_i$. The obtained hierarchical model is equivalent to the one presented in Section 3, with unchanged parameters $\theta = (\overline{y_0}, \overline{c_0}, \overline{m_0}, \overline{A}, t_0, \sigma_\tau^2, \sigma_\xi^2, \sigma_\epsilon^2)$ and equivalent random effects $z = (z_{\text{pop}}, z_1, ..., z_N)$, where $z_{\text{pop}} = (y_0, c_0, m_0, A)$ and $\forall i \in [\![1, N]\!]$, $z_i = (t_i, \xi_i, s_i)$. The complete log-likelihood writes:

$$\log q(y, z, \theta) = \sum_{i=1}^N \sum_{j=1}^{n_i} \log q(y_{i,j}|z, \theta) + \log q(z_{\text{pop}}|\theta) + \sum_{i=1}^N \log q(z_i|\theta) + \log q(\theta) \quad (6)$$

where the densities $q(y_{i,j}|z, \theta)$, $q(z_{\text{pop}}|\theta)$, $q(z_i|\theta)$ and $q(\theta)$ are given, up to an additive constant, by:

$$-2 \log q(y_{i,j}|z, \theta) \stackrel{+\text{cst}}{=} \Lambda \log \sigma_\epsilon^2 + \|y_{i,j} - \eta_{c_0, m_0, t_0, \psi_i(t_{i,j})}(w_i) \circ y_0\|^2 / \sigma_\epsilon^2 \quad (7)$$

$$-2 \log q(z_{\text{pop}}|\theta) \stackrel{+\text{cst}}{=} |y_0| \log \sigma_y^2 + \|y_0 - \overline{y_0}\|^2/\sigma_y^2 + |c_0| \log \sigma_c^2 + \|c_0 - \overline{c_0}\|^2/\sigma_c^2 \quad (8)$$
$$+ |m_0| \log \sigma_m^2 + \|m_0 - \overline{m_0}\|^2/\sigma_m^2 + |A| \log \sigma_A^2 + \|A - \overline{A}\|^2/\sigma_A^2$$

$$-2 \log q(z_i|\theta) \stackrel{+\text{cst}}{=} \log \sigma_\tau^2 + (t_i - t_0)^2/\sigma_\tau^2 + \log \sigma_\xi^2 + \xi_i^2/\sigma_\xi^2 + \|s_i\|^2 \quad (9)$$

$$-2 \log q(\theta) \stackrel{+\text{cst}}{=} \|\overline{y_0} - \overline{\overline{y_0}}\|^2/\varsigma_y^2 + \|\overline{c_0} - \overline{\overline{c_0}}\|^2/\varsigma_c^2 + \|\overline{m_0} - \overline{\overline{m_0}}\|^2/\varsigma_m^2 + \|\overline{A} - \overline{\overline{A}}\|^2/\varsigma_A^2 \quad (10)$$
$$+ (t_0 - \overline{\overline{t_0}})^2/\varsigma_t^2 + m_\tau \log \sigma_\tau^2 + m_\tau \sigma_{\tau,0}^2/\sigma_\tau^2 + m_\xi \log \sigma_\xi^2 + m_\xi \sigma_{\xi,0}^2/\sigma_\xi^2$$
$$+ m_\epsilon \log \sigma_\epsilon^2 + m_\epsilon \sigma_{\epsilon,0}^2/\sigma_\epsilon^2$$

noting $\Lambda$ the dimension of the space where the residual $\|y_{i,j} - \eta_{c_0, m_0, t_0, \psi_i(t_{i,j})}(w_i) \circ y_0\|^2$ is computed, and $|y_0|, |c_0|, |m_0|$ and $|A|$ the total dimension of $y_0$, $c_0$, $m_0$ and $A$ respectively. We chose either the current [42] or the varifold [7] norm for the residuals.

Noticing the identity $\eta_{c_0, m_0, t_0, \psi_i(t_{i,j})} = \eta_{c_0, m_0, 0, \psi_i(t_{i,j}) - t_0}$, the complete log-likelihood can be decomposed into $\log q(y, z, \theta) = \langle S(y, z), \Phi(\theta) \rangle_{\text{Id}} - \Psi(\theta)$ i.e. the proposed mixed-effects model belongs the curved exponential family. In this setting, the MCMC-SAEM algorithm presented in Section 4 has a proved convergence.

Exhibiting the sufficient statistics $S_1 = y_0$, $S_2 = c_0$, $S_3 = m_0$, $S_4 = A$, $S_5 = \sum_i t_i$, $S_6 = \sum_i t_i^2$, $S_7 = \sum_i \xi_i^2$ and $S_8 = \sum_i \sum_j \|y_{i,j} - \eta_{c_0, m_0, t_0, \psi_i(t_{i,j})}(w_i) \circ y_0\|^2$ (see Section 4.5), the update of the model parameters $\theta \leftarrow \theta^\star$ in the M step of the MCMC-SAEM algorithm can be derived in closed form:

$$\overline{y_0}^\star = [\varsigma_y^2 S_1 + \sigma_y^2 \overline{\overline{y_0}}] / [\varsigma_y^2 + \sigma_y^2] \qquad t_0^\star = [\varsigma_t^2 S_5 + \sigma_\tau^{2\star} \overline{\overline{t_0}}] / [N\varsigma_t^2 + \sigma_\tau^{2\star}] \quad (11)$$

$$\overline{c_0}^\star = [\varsigma_c^2 S_2 + \sigma_c^2 \overline{\overline{c_0}}] / [\varsigma_c^2 + \sigma_c^2] \qquad \sigma_\tau^{2\star} = [S_6 - 2t_0^\star S_5 + N t_0^{\star 2} + m_\tau \sigma_{\tau,0}^2] / [N + m_\tau] \quad (12)$$

$$\overline{m_0}^\star = [\varsigma_m^2 S_3 + \sigma_m^2 \overline{\overline{m_0}}] / [\varsigma_m^2 + \sigma_m^2] \qquad \sigma_\xi^{2\star} = [S_7 + m_\xi \sigma_{\xi,0}^2] / [N + m_\xi] \quad (13)$$

$$\overline{A}^\star = [\varsigma_A^2 S_4 + \sigma_A^2 \overline{\overline{A}}] / [\varsigma_A^2 + \sigma_A^2] \qquad \sigma_\epsilon^{2\star} = [S_8 + m_\epsilon \sigma_{\epsilon,0}^2] / [\Lambda N \langle n_i \rangle_i + m_\epsilon] \quad (14)$$

The intricate update of the parameters $t_0 \leftarrow t_0^\star$ and $\sigma_\tau^2 \leftarrow \sigma_\tau^{2\star}$ can be solved by iterative replacement.

Similarly to Equation 6, the tempered complete log-likelihood writes:

$$\log q_T(y, z, \theta) = \sum_{i=1}^N \sum_{j=1}^{n_i} \log q_T(y_{i,j}|z, \theta) + \log q_T(z_{\text{pop}}|\theta) + \sum_{i=1}^N \log q(z_i|\theta) + \log q(\theta) \quad (15)$$

with:
$$-2 \log q_T(y_{i,j}|z, \theta) \stackrel{+\text{cst}}{=} \Lambda \log(T\sigma_\epsilon^2) + \|y_{i,j} - \eta_{c_0, m_0, t_0, \psi_i(t_{i,j})}(w_i) \circ y_0\|^2/(T\sigma_\epsilon^2) \quad (16)$$

$$-2 \log q_T(z_{\text{pop}}|\theta) \stackrel{+\text{cst}}{=} |y_0| \log(T\sigma_y^2) + \|y_0 - \overline{y_0}\|^2/(T\sigma_y^2) + |c_0| \log(T\sigma_c^2) + \|c_0 - \overline{c_0}\|^2/(T\sigma_c^2) \quad (17)$$
$$+ |m_0| \log(T\sigma_m^2) + \|m_0 - \overline{m_0}\|^2/(T\sigma_m^2) + |A| \log(T\sigma_A^2) + \|A - \overline{A}\|^2/(T\sigma_A^2)$$

Tempering can therefore be understood as an artificial increase of the variances $\sigma_\epsilon^2$, $\sigma_y^2$, $\sigma_c^2$, $\sigma_m^2$ and $\sigma_A^2$ when computing the associated acceptation ratios in the S-MCMC step of the algorithm. This intuition is well-explained in [25].